\documentclass[runningheads]{llncs}
\usepackage[T1]{fontenc}
\usepackage{graphicx}
\usepackage{amsmath}
\usepackage{amssymb}
\usepackage{booktabs}
\usepackage{wrapfig}
\usepackage{xcolor}
\usepackage{ulem}
\usepackage[misc]{ifsym}

\usepackage{tikz}
\usepackage{float}
\usetikzlibrary{positioning, arrows.meta, shapes.geometric}
\usepackage{wrapfig}
\usepackage{orcidlink}
\begin{document}
\title{Motion Primitive Discovery in a Humanoid Robot via Self-Organising Maps for Phase Recognition}
\titlerunning{Motion Primitive Discovery for Online Phase Recognition}
\author{Radovan Gregor \and
Igor Farka\v{s}\textsuperscript{\orcidlink{0000-0003-3503-2080}
}}
\authorrunning{R. Gregor and I. Farka\v{s}}
%
\institute{Department of Applied Informatics\\ Comenius University Bratislava, Slovakia \\ 
\email{\{radovan.gregor,igor.farkas\}@fmph.uniba.sk} 
}
\maketitle

\makeatletter
\def\convertto#1#2{\strip@pt\dimexpr #2*65536/\number\dimexpr 1#1}
\makeatother

\begin{abstract}
Understanding the computational basis of action recognition is a central challenge in social cognition as well as in human--robot interaction. Inspired by the Mirror Neuron System (MNS), we propose a two-level architecture for motor primitive discovery and online phase recognition applied to the NICO humanoid robot. At the first level, two Self-Organising Maps (SOMs) learn topographic representations of arm kinematics (A-SOM) and hand kinematics (H-SOM) from simulated trials covering seven motor actions. The maps are trained on non-redundant features identified through hierarchical correlation analysis of motion trajectories.
The results show that the two SOMs encode complementary aspects of motor behaviour.
At the second level, an Echo State Network (ESN) evaluates whether temporal trajectories of SOM activations, represented by consecutive best-matching units, are sufficient for online recognition of the currently executed movement phase. 
The results show that SOM-based trajectories preserve the dominant phase-discriminative structure of the movement, while contextual information provides only a secondary refinement.  
Our contribution is the integration of established SOM and ESN methods within an MNS-inspired architecture for motor primitive representation and online phase recognition. The results are compatible with the computational hypothesis that self-organised motor representations, when temporally integrated, can support accurate online recognition of ongoing movement phases. 

\keywords{Motion Primitive \and Phase Recognition \and SOM \and ESN}
\end{abstract}

\section{Introduction and Related Work}
\label{sec:intro} 

The ability to recognise observed actions and infer the intentions behind them is fundamental to human--robot interaction \cite{Thomaz2013}. A robot that can interpret a human partner's motor behaviour can proactively adapt, anticipate, and collaborate more effectively. 
The Mirror Neuron System (MNS) offers a biological model for this capacity: mirror neurons discharge both when an individual executes a motor act and when the same individual observes another performing the same or a similar motor act \cite{Gallese1996}, suggesting that action understanding is grounded in an internal motor vocabulary rather than abstract perceptual classification \cite{Rizzolatti2004}. 
Neurophysiological studies have shown that parietal mirror neurons encode not only the observed action but also its goal \cite{Fogassi2005}.
This is a key feature of the action-understanding hypothesis and motivates the design of computational systems that can similarly distinguish actions at the level of motor primitives and at the level of goal-directed sequences \cite{Oztop2006}.
 
Modern approaches encode observed motion directly from kinematic or visual signals using recurrent or transformer architectures trained end-to-end on action labels \cite{Bertasius2021}. 
These approaches do not always explicitly emphasise the biological insight that action understanding operates across two levels simultaneously: a repertoire of elementary motion primitives and the goal-directed sequences they compose \cite{Wolpert2003}.
Computational models of the MNS have addressed this by modelling the mirror system as a Self-Organising Map (SOM) \cite{Kohonen1982} that
receives both motion and context inputs, showing that goal-specific neuron pools emerge naturally from the geometric relationships between these inputs without any explicit goal representation being programmed in \cite{Thill2011}. The same principle underpins chain models of intention understanding \cite{Chersi2011,Erlhagen2006}, in which sequences of motor primitives can be linked over time to recognise or predict higher-level actions. 
In our previous MNS-related work, we also exploited topographic mapping as an organising principle \cite{Pospich19,Rebrova13}.
Here we focus on topographic representation of proprioceptive signals whose sequence describes the time-varying state of moving arm and hand. Proprioception is one of the least understood senses in cognitive neuroscience, yet fundamental for the control of movement. There is evidence that cortical topography is a widespread organising principle of the brain \cite{deCharms2000}. 

We propose a two-level MNS-inspired architecture drawing on these findings. At the lower level, two SOMs discover reusable motion primitives from arm and hand kinematics, 
consistent with the observation that motion and context inputs to parietal mirror neurons originate from distinct neural pathways \cite{Thill2011}. 
The topographic projection groups temporally similar movements into prototype nodes, which serve as a discrete primitive vocabulary \cite{Hemeren2011}. This dual-map design reflects the fact that arm movement and hand shaping contribute complementary information during action execution. 
At the upper level, an Echo State Network (ESN) \cite{Luko09} encodes the temporal ordering of these primitives into a reservoir state from which movement phases can be inferred, extending our previous ESN-based approach to arm trajectory prediction \cite{Gregor2025China}. 

In this paper, we focus on the evaluation of the second level, in which an ESN-based temporal readout tests whether SOM trajectories are sufficient for online recognition of the currently executed movement phase. 
The objective is not to classify a complete trial after observing the whole sequence, but to recognise the current movement phase online from the sensory-motor stream available up to that timestep. This corresponds to a classification problem in which the reservoir state encoding recent SOM activations is used to predict the current phase.
Our main research question is: How much of the robot's current movement-phase structure is determined by the temporal dynamics of learned motor primitives alone, and how much additional contextual information is required for accurate online phase recognition? The study tests whether the SOM representation is sufficient, not whether it outperforms a direct-kinematics ESN or modern sequence-learning models.
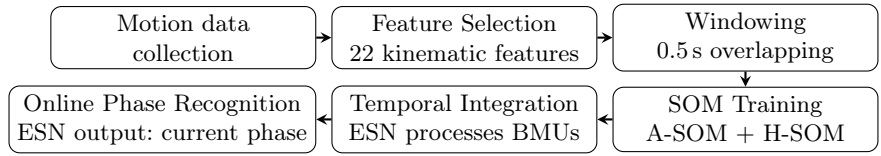
\begin{figure}[h!]
\centering
\begin{tikzpicture}[
    node distance=0.2cm,
    >=stealth,
    process/.style={
        rectangle,
        rounded corners,
        draw,
        align=center,
        minimum width=3.5cm,
        minimum height=0.8cm,
        font=\small
    },
    data/.style={
    rectangle,
    draw,
    dashed,
    align=center,
    minimum width=3.5cm,
    minimum height=0.9cm,
    font=\small
    },
    arrow/.style={->, thick}
]

\node[process] (s1) {Motion data \\ collection};
\node[process, right=of s1] (s2) {Feature Selection\\22 kinematic features};
\node[process, right=of s2] (s3) {Windowing\\0.5\,s overlapping };
\node[process, below=of s3] (s4) {SOM Training\\A-SOM $+$ H-SOM};
\node[process, left=of s4] (s5) {Temporal Integration\\ESN processes BMUs};
\node[process, left=of s5] (s6) {Online Phase Recognition\\ESN output: current phase};

\draw[arrow] (s1) -- (s2);
\draw[arrow] (s2) -- (s3);
\draw[arrow] (s3) -- (s4);
\draw[arrow] (s4) -- (s5);
\draw[arrow] (s5) -- (s6);

\end{tikzpicture}
\caption{Processing pipeline of the SOM+ESN motion primitive recognition
system.}
\label{fig:pipeline}
\end{figure}
\section{Methods}
\label{sec:methods}

Here we describe the whole data processing pipeline illustrated in Fig.~\ref{fig:pipeline}.

\subsection{Data Collection in Simulation}

To build the dataset for motion primitive discovery, we employ a physics-based Unity simulation environment featuring the NICO robot \cite{Kerzel2017}, a humanoid platform for multimodal interaction. NICO performs seven motor actions -- \textit{Pick, Eat, Place, Push, Tap, Point, Wave}. Each action is performed on a library of 14 physically diverse objects spanning spheres, cubes, cylinders, discs and irregular shapes in small, medium and large sizes. 

\begin{wraptable}{r}{0.4\textwidth}
  \centering
  \vspace{-30pt}
  \caption{Feature groups.}
  \label{tab:columns}
  \begin{tabular}{lc}
  \hline
  \textbf{Group} & \textbf{Count} \\ \hline
  Metadata / labels     & 6  \\
  Arm joint angles      & 6  \\
  Arm joint velocities  & 6  \\
  TCP positions         & 15 \\
  TCP velocities        & 9  \\
  Palm pose             & 7  \\
  Finger joints         & 39 \\
  Aperture              & 4  \\
  Object state          & 16 \\
  \textbf{Total}        & \textbf{108} \\ \hline
  \end{tabular}
  \vspace{-20pt}
\end{wraptable}

\begin{figure}[t!]
	\centering
	\includegraphics[width=0.9\columnwidth]{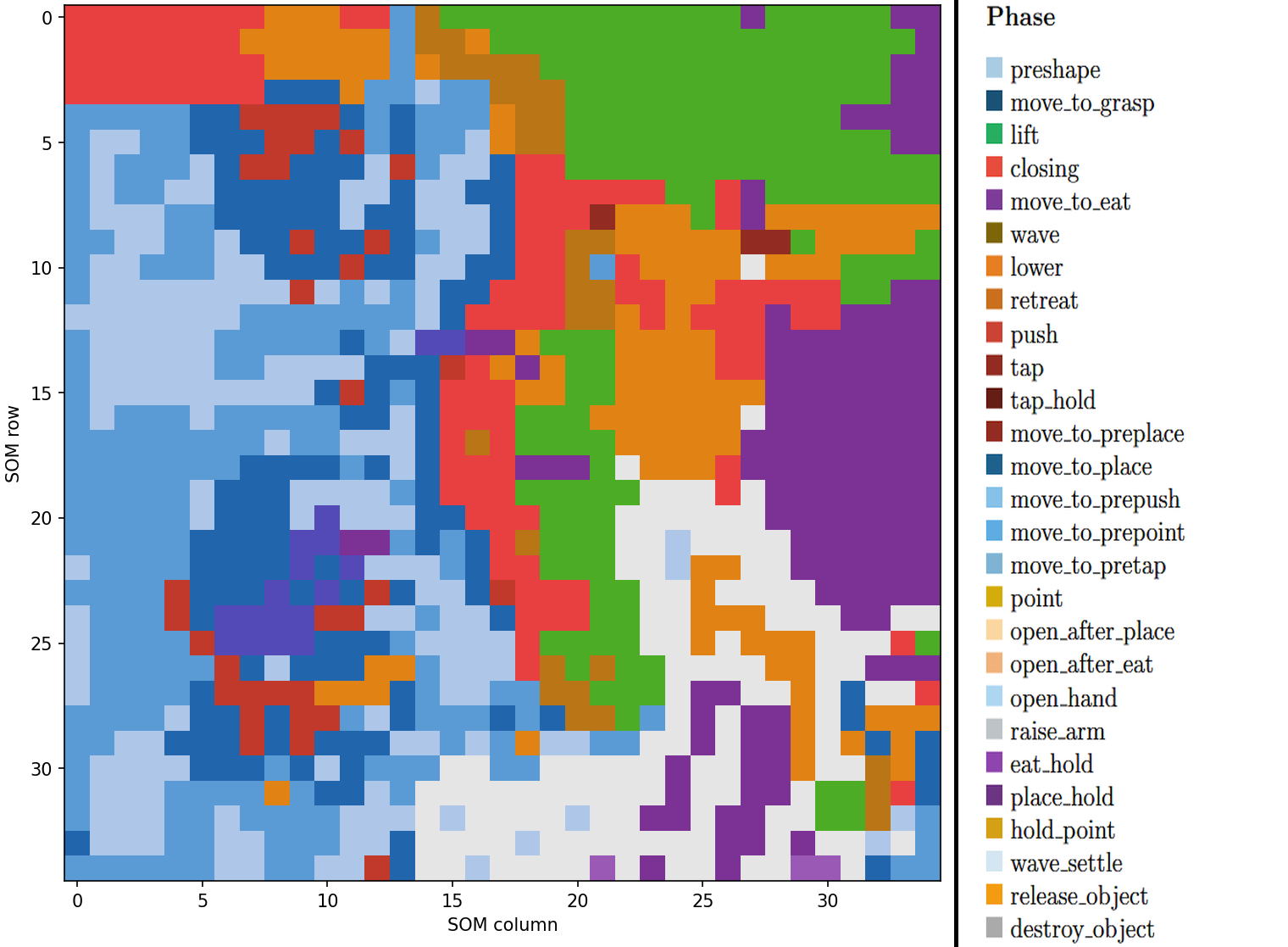}
	\caption{A-SOM dominant phase per neuron. Neuron colour indicates the phase whose training samples most frequently activated that neuron. The map exhibits clear topographical organisation (Mean
	Phase Purity 73.9\%).}\label{fig:a-som_phase}
\end{figure}

Object-manipulation actions (\textit{Pick, Eat, Place}) are executed with either a Power or Precision grasp, while the remaining actions each employ a dedicated hand configuration. 
To ensure diversity in the dataset, object positions were randomized within a predefined reachable workspace (peripersonal space), requiring the robot to generate varying motion trajectories.
A Barracuda ONNX IK network, which is part of the simulation platform, maps Cartesian poses of the desired Tool Center Point to 6 joint angles.
The 13 finger joints of the hand are controlled by four scalar parameters -- thumb rotation, thumb curl, index closure, and finger closure -- each rigidly driving a fixed anatomical joint group, with the joints within each group moving in fixed proportions by design. This low-dimensional synergy controller is grounded in human motor neuroscience \cite{Santello1998}.
Motion is logged at 60 Hz, corresponding to one frame every approximately 16.7 ms. Each frame contains 108 simultaneously recorded variables, and the complete dataset comprises 4,200 trials. As described in Section~\ref{sec:feature-selection}, not all 108 features are used as model inputs.
The recorded variables are organised into nine functional groups as shown in Table~\ref{tab:columns}.
In this dataset, an action denotes a complete goal-directed trial. Seven actions are implemented as phase-sequenced state machines, each decomposed into a sequence of named movement phases.
For example, the \textit{Eat} action follows the sequence \textit{preshape} $\rightarrow$ \textit{move-to-grasp} $\rightarrow$ \textit{closing} $\rightarrow$ \textit{move-to-eat} $\rightarrow$ \textit{eat-hold} $\rightarrow$ \textit{open-after-eat}. Across all seven actions, this yields 27 distinct phase labels, all of which are enumerated in the legend of Figure~\ref{fig:a-som_phase}.
A motion primitive is more fundamental. 
It is a compact pattern of movement that the SOM discovers from the kinematic data. It does not belong to one action or one phase but it is a reusable building block. One phase can contain multiple primitives while one primitive can appear across multiple actions.

\subsection{Feature Analysis and Selection}
\label{sec:feature-selection}
Prior to SOM training, a three-stage analysis pipeline was applied to the 108 candidate features to identify and remove redundancy, verify cluster structure in window space, and confirm feature independence. In Stage 1, 51 columns are excluded on functional grounds: metadata and trial labels that carry no kinematic information, absolute joint positions that are workspace-dependent and do not generalise across object locations, and duplicated TCPs. This leaves 57 candidate kinematic features for correlation analysis.

Next we quantify how strongly each pair of features covaries across all $N$ filtered frames. This relationship is first captured by \textit{covariance}. Here, $N$ denotes the \textit{number of rows} (frames) in the filtered dataset -- 1,284,794 frames across 4,200 trials, with 306 frames per trial on average. The covariance, however, depends on the physical units of each feature -- joint velocities in deg/s and aperture in meters produce incomparable magnitudes. The Pearson correlation coefficient removes this scale dependence by dividing the covariance by the standard deviation of each feature. Taking the absolute value $|r_{ij}|$ treats positive and negative correlations equally, since both indicate redundancy. 
The result is a symmetric similarity matrix $\mathbf{R} \in [0,1]^{D \times D}$ with $r_{ii} = 1$, where $D = 57$ is the number of candidate features. Computing the correlation on raw frames (not windowed) yields the covariance structure unaffected by window size choice.

$\mathbf{R}$ is converted to a distance matrix by a simple transformation $d_{ij} = 1 - r_{ij},$ \rm{where} $d_{ij} \in [0, 1]$.
Hierarchical agglomerative clustering is applied to the condensed distance matrix using Ward's minimum-variance linkage. The linkage matrix $\mathbf{Z}$ produced by this process encodes the full merge tree and is visualised as a dendrogram. 
Features that branch at low distance (high $r_{ij}$) are considered strongly correlated; only one representative from each such cluster is retained for SOM training. Cutting the dendrogram at $\tau = 0.7$ produced 22 non-redundant features from the original 57 candidates.  The final set of 22 non-redundant features is divided into: (1)~13 arm features for the arm SOM (A-SOM) -- TCP velocities (3), arm joint velocities (6), palm orientation quaternion (4); and (2)~9 hand features for the hand SOM (H-SOM) -- aperture and velocity (2), thumb and middle finger positions (3), finger joint velocities (4).

\subsection{Windowing and Normalisation}

Motion logs were segmented into overlapping windows of $W=30$ frames (0.5\,s at 60\,Hz) with stride $S=15$ frames (50\% overlap).
Each window is represented by its per-feature mean; phase and action labels are taken from the central frame. Actions are stratified to equal window counts to prevent over-represented actions from dominating SOM node placement, yielding 8,789 windows per action (61,523 total). 
Each feature is independently normalised using $z$-score standardisation computed from training-set statistics, yielding vectors 
$\hat{\mathbf{x}}(t)$.
Normalisation is applied independently for A-SOM and H-SOM, as an
essential step because features span incompatible units and scales.

\section{The SOM models}
\label{sec:som}
\subsection{The model training}
Each SOM (with a grid $R$$\times$$C = K$ neurons) learns a weight matrix $\mathbf{W} \in \mathbb{R}^{K \times d}$, where $d$ is the SOM-specific input dimension ($d_A = 13$ for A-SOM, $d_H = 9$ for H-SOM). The best matching unit (BMU) for input $\hat{\mathbf{x}}(t)$ is the neuron $i^{*}$ minimising the Euclidean distance between input vector and BMU weight vector.
During training, weights are updated according to the Kohonen rule.
For training step $n$ with learning rate $\alpha(n)$ and neighbourhood
function $h_{i,i^{*}}(t)$:
\begin{equation}\label{eq:update}
  \mathbf{w}_{i}(t{+}1) = \mathbf{w}_{i}(t)
  + \alpha(t)\, h_{i^{*}i}(t)
  \left[ \hat{\mathbf{x}}(t) - \mathbf{w}_i(t) \right]
\end{equation}
where $h_{i^{*},i}(t)$ is the Gaussian neighbourhood function shrinking over time.

To quantify map organisation, we measured the purity of phase, action and grasp. For each active neuron, the dominant class label was identified. Purity was then computed as the proportion of samples belonging to that dominant label, averaged across active neurons. Phase purity measures the consistency of movement-phase representation. Neuron utilization was computed as the percentage of neurons that became BMUs for at least one training sample.

\begin{wraptable}{r}{0.55\textwidth}
\centering
  \vspace{-25pt}
\caption{Selected SOM configurations \\ and purity metrics.}
\label{tab:results}
\small
\begin{tabular}{lccccc}
\toprule
\textbf{SOM} & \textbf{Grid} & \textbf{Active} &
\textbf{Phase} & \textbf{Action} & \textbf{Grasp} \\
\midrule
Arm  & 35$\times$35 & 90.0\% & 73.9\% & 66.9\% & 66.4\% \\
Hand & 18$\times$18 & 51.5\% & 76.1\% & 72.9\% & 97.2\% \\
\bottomrule
\end{tabular}
  \vspace{-20pt}
\end{wraptable}

A-SOM uses a 35$\times$35 grid of neurons, justified by a systematic grid search that showed that increasing the grid size from 25$\times$25 improved mean phase purity by $+5.5$ percentage points (pp), an absolute difference between two percentages, while maintaining $90\%$ active neurons. 
H-SOM uses an 18$\times$18 grid. 
Testing showed that the hand configuration space is intrinsically low-dimensional because the synergy controller produces approximately six distinct hand poses, limiting the effective vocabulary size. The selected configurations are summarised in Table~\ref{tab:results}.

\subsection{Results}
 
A-SOM achieves 90.0\% neuron utilisation, with dead neurons concentrating at the map periphery as expected. Phase purity (73.9\%) is the strongest label dimension, with large coherent zones emerging without supervision:
\textit{lift} anchors the top-right corner, \textit{wave} occupies the right edge, and \textit{preshape} spans the centre-left, transitioning smoothly into \textit{move-to-grasp} --- reflecting the temporal order of a reach--grasp cycle. The map encodes a directional reaching topology: approach phases occupy the left half and terminal or post-action phases the right, a gradient learned entirely from kinematics with no temporal supervision. 
Action purity (66.9\%) and grasp purity (66.4\%) are lower; arm kinematics alone cannot distinguish actions sharing identical reach trajectories, nor encode grip type, both of which are delegated to H-SOM.
 
\begin{figure}[t!]
	\centering
    \includegraphics[width=0.9\columnwidth]{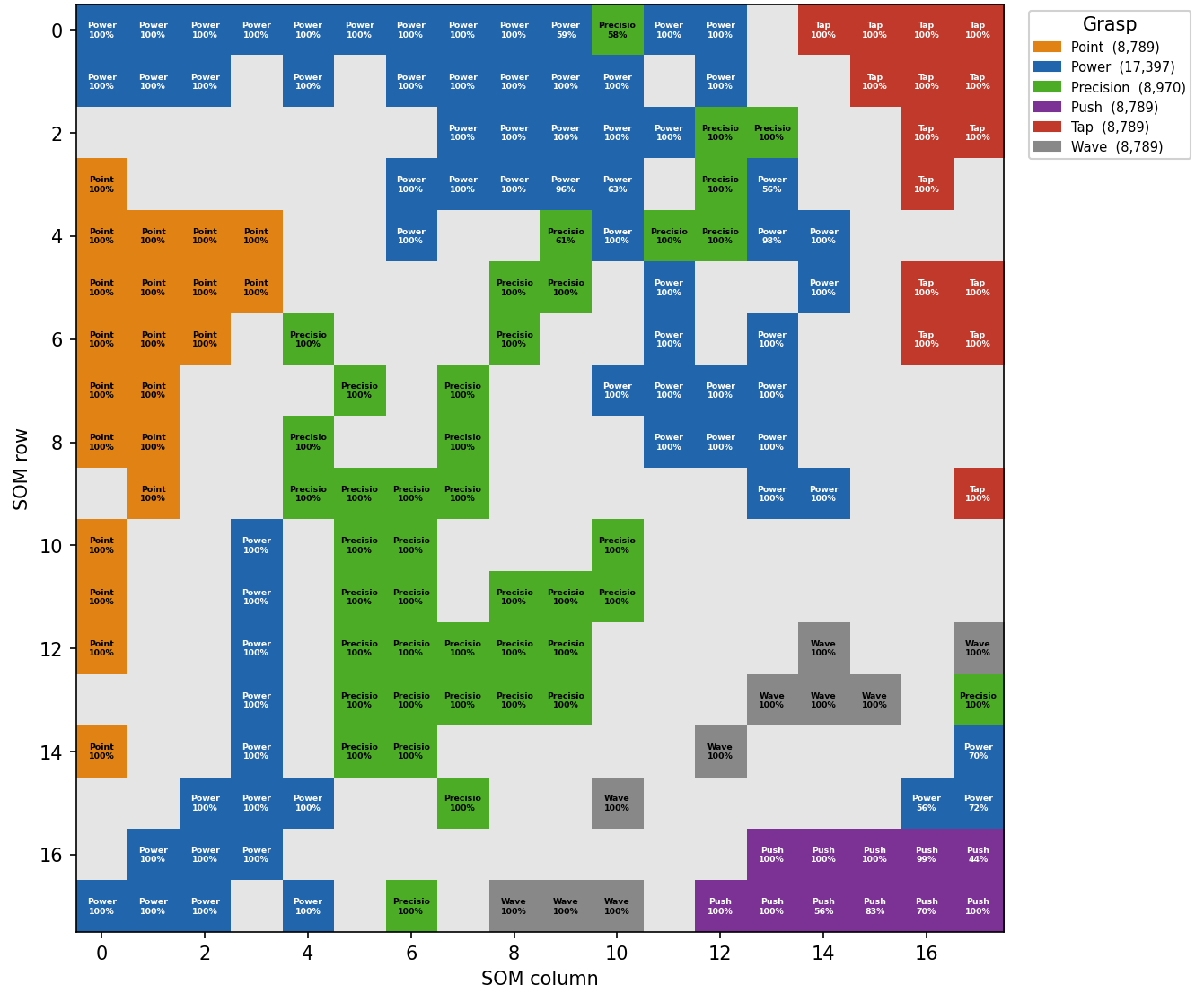}
	\caption{H-SOM dominant grasp type per neuron. Neuron colour indicates the dominant grasp synergy with its purity percentage. 
	(Mean Purity 97.2\%).}\label{fig:h-som_grasp}
\end{figure}

H-SOM achieves 51.5\% neuron utilisation, which is lower than that of A-SOM by design -- the hand synergy controller produces discrete, non-overlapping configuration spaces, so dead neurons act as
hard boundaries between isolated action islands rather than representing wasted capacity. Grasp purity (97.2\%)  (Fig.~\ref{fig:h-som_grasp}) is the headline result: \textit{Point, Tap, Wave}, and \textit{Push} each form perfectly pure, isolated clusters (100\% throughout), while Power and Precision separate into non-overlapping zones with impurities confined to a narrow mid-close transition band.
Importantly, the map encodes a \textit{configuration topology} rather than a temporal one -- each hand pose occupies its own island with no kinematic path between them, reflecting the hard discreteness of the
synergy controller. Action purity (72.9\%) and phase purity (76.1\%, exceeding A-SOM) confirm that finger-level signals carry complementary discriminative structure to the arm-level primitives of A-SOM.
 
\begin{figure}[h!]
	\centering
	\includegraphics[width=0.8\columnwidth]{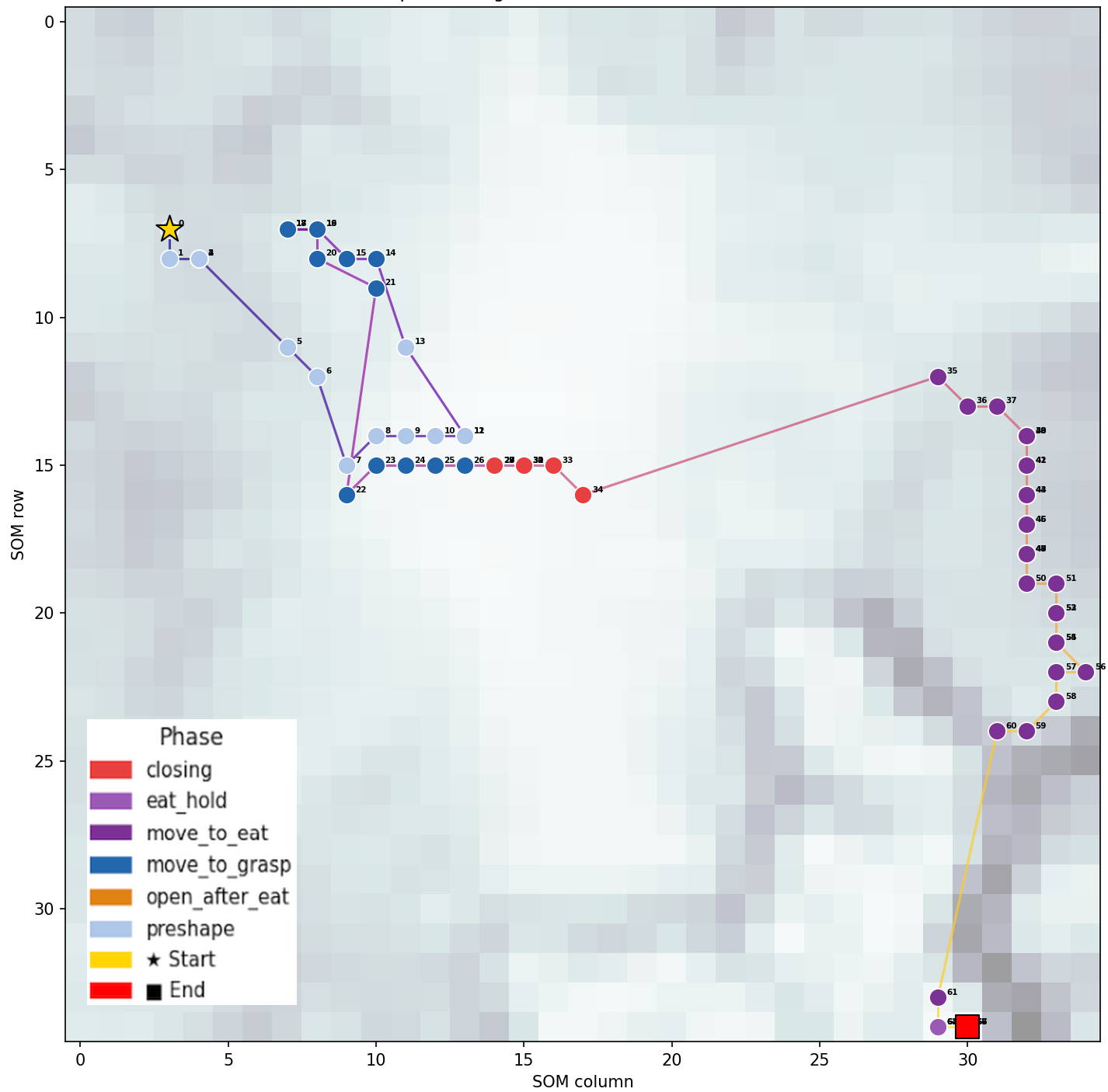}
	\caption{Single-trial trajectory on A-SOM representing \textit{Eat} action traces a smooth left-to-right path. Across 68 timesteps, the trajectory visits 46 unique neurons, indicating that the movement progresses through a structured sequence of arm-motion primitives. }\label{fig:bmu_trial}
\end{figure}
Figure~\ref{fig:bmu_trial} visualises the trajectory of BMUs activated on the A-SOM during a single trial. Each point corresponds to the winning SOM neuron at one timestep, and the numbered markers indicate the temporal order of the trajectory. 
The path shows how the arm movement evolves across the SOM surface as the action progresses through different phases, from the start marker to the final endpoint. Colours denote the ground-truth movement phase at each timestep, illustrating that consecutive phases occupy coherent regions of the arm map and that the SOM trajectory preserves the temporal structure of the executed action.

\section{Phase Recognition using the ESN}
\label{sec:phaserecog}

\subsection{Research Questions}

The dual-SOM architecture described in Section~\ref{sec:som} produces, for each trial timestep, a pair of BMU weight vectors that jointly encode the current arm kinematic state and hand configuration. 
To evaluate whether this representation is sufficient for online phase recognition, the temporal sequence of these primitive descriptors is integrated by the ESN.
Given the reservoir state $\mathbf{x}(t)$, accumulated from the trajectory of SOM activations up to time $t$, the task is to predict the phase label $\phi(t)$ active at that moment.

We address two sub-questions: 1. How accurately can the ESN identify the current movement phase from SOM activation trajectories alone, without any contextual information? 2. How much does additional  information -- action identity, object distance, contact state -- contribute beyond the primitive signal?

\subsection{Experimental Setup}

\subsubsection{Dataset and Split}

The dataset consists of the original $4{,}200$ simulated trials from SOM training, subsampled at stride $S=5$ (83\,ms per step at 60\,Hz). A stratified 80:20 split by action yields $3{,}360$ training trials and 840 test trials. After removing the warm-up period of $W=5$ steps per trial, the training set contains $190{,}603$ labelled timesteps.

\subsubsection{Input Representation}
For each timestep $t$, the raw kinematic features are first $z$-score normalised using training-set statistics computed independently per SOM.
The BMU is then identified and the input to the ESN is constructed from the BMU weight vectors on A-SOM and H-SOM.
Specifically, we get $\mathbf{b}^{\mathrm{A}}(t) = \mathbf{w}^{\rm A}_{i^{*}} \in \mathbb{R}^{13}$
and
$\mathbf{b}^{\mathrm{H}}(t) = \mathbf{w}^{\rm H}_{i^{*}} \in \mathbb{R}^{9}$.

Four context configurations, accumulatively extended, are evaluated:
\begin{enumerate}
  \item \textit{ctx-no} (22D): primitive descriptor only,
        $\mathbf{u}(t) = [\mathbf{b}^{\mathrm{A}}(t),\;
        \mathbf{b}^{\mathrm{H}}(t)]$
  \item \textit{ctx-act} (29D): + action one-hot
        $\mathbf{a} \in \{0,1\}^7$
  \item \textit{ctx-dst} (30D): + normalised palm--object distance $d(t) \in \mathbb{R}$
  \item \textit{ctx-all} (31D): + binary object contact flag $c(t) \in \{0,1\}$
\end{enumerate}

\subsubsection{Target}
The readout target at each timestep is the one-hot encoding of the ground-truth phase $\phi(t)$ currently active:
\begin{equation}\label{eq:phasetarget}
  \mathbf{y}(t) = \mathbf{e}_{\phi(t)} \in \{0,1\}^{27}.
\end{equation}

\subsubsection{Reservoir}
The ESN reservoir consists of $N=300$ randomly connected neurons.
The state update is given by:
\begin{align}
  \label{eq:esn}
  \tilde {\mathbf{x}}(t) &= {\rm tanh} (\mathbf{W}^{\text{res}} \mathbf{x}(t-1) + \mathbf{W}^{\text{in}} \mathbf{u}(t)) \\
  \mathbf{x}(t) &= (1 - \alpha)\mathbf{x}(t-1) + \alpha \tilde{\mathbf{x}}(t),
\end{align}
where $\mathbf{x}(t) $ is the reservoir state at time $t$, $\tilde {\mathbf{x}}(t)$ is an internal state variable, and leaking rate $\alpha=0.50$.
Weight matrices $\mathbf{W}^{\rm in}$ and $\mathbf{W}^{\rm res}$ are randomly initialised (the reservoir matrix scaled to spectral radius $\rho=0.99$) and remain fixed throughout training; only the readout $\mathbf{W}^{\rm out}$ is trained. 

\subsubsection{Readout Training}
Reservoir states are collected during a single teacher-forced pass through all training trials. The reservoir state is reset to zero at the start of each trial, ensuring that no information leaks across trials. The first $W = 5$ steps serve as warm-up and are excluded from state collection, yielding the state matrix $\mathbf{X} \in \mathbb{R}^{M \times N}$ ($M = 190{,}603$ post-warm-up steps, $N = 300$ reservoir neurons).
The readout weights are computed analytically via ridge regression:
\begin{equation}\label{eq:phaseridge2}
  \mathbf{W}^{\rm out} =
  \left(\mathbf{X}^\top\mathbf{X}
        + \lambda\,\mathbf{I}\right)^{-1}
  \mathbf{X}^\top\mathbf{Y}, \quad \lambda = 10^{-4}
\end{equation}
where $\mathbf{Y} \in \mathbb{R}^{M \times 27}$ is the matrix of current-phase one-hot targets. 

\subsubsection{Inference}
At each test step $t$, the predicted phase is
\begin{equation}\label{eq:phaseinfer2}
  \hat{\phi}(t) =
  \arg\max_{k}\,
  \left[\mathbf{W}_{\mathrm{out}}^\top\,\mathbf{x}(t)\right]_k
\end{equation}
The system operates as an online classifier: the reservoir state $\mathbf{x}(t)$ is driven by the ground-truth input $\mathbf{u}(t)$ at each step, the readout produces a
phase estimate without generating future inputs.
Phase recognition accuracy is evaluated as the fraction of post-warm-up steps where the predicted phase matches ground-truth phase.

\section{ESN Results}

\begin{wraptable}{r}{0.35\textwidth}
\centering
\vspace{-11mm}
\caption{PRA (in \%) across context configurations.
 $N$=300, $\rho$=0.99, $\alpha$=0.50.}
\label{tab:phase_recog}
\small
\begin{tabular}{lccc}
\toprule
\textbf{Config} & \textbf{Dims} & \textbf{PRA} & \textbf{Gain} \\
\midrule
ctx-no  & 22D & 93.9  & ---        \\
ctx-act & 29D & 94.6  & +0.7\,pp  \\
ctx-dst & 30D & 94.6  & +0.8\,pp  \\
ctx-all & 31D & \textbf{94.9} & +1.0\,pp  \\
\bottomrule
\end{tabular}
  \vspace{-20pt}
\end{wraptable}

\subsection{Phase Recognition Accuracy}
Table~\ref{tab:phase_recog} reports overall phase recognition accuracy (PRA) across the four context configurations. All results are reported on the held-out test set. The primitive descriptor alone (\textit{ctx-no}) achieves $93.9\%$, indicating that SOM activation trajectories carry the dominant share of phase information without any world-state context.
The full context configuration (\textit{ctx-all}) adds $+1.0$\,pp over the no-context baseline, confirming that contextual signals play a secondary role once the temporal integration of primitive descriptors has been performed by the
reservoir.

\subsection{Hyperparameter Sensitivity}

\begin{wraptable}{r}{0.65\textwidth}
\centering
\vspace{-11mm}
\caption{PRA (\%) for selected reservoir configurations (9 run grid omitted for brevity).}
\label{tab:hps}
\small
\begin{tabular}{lccccccccc}
\toprule
\textbf{Run} & $N$ & $\rho$ & $\alpha$ & $\lambda$ &
ctx-no & ctx-act & ctx-dst & ctx-all \\
\midrule
R1 & 30   & 0.99 & 0.50 & $10^{-4}$ & 73.6 & 79.0 & 75.5 & 77.7 \\
R5 & 300  & 0.99 & 0.50 & $10^{-4}$ & 93.9 & 94.6 & 94.6 & 94.9 \\
R8 & 500  & 0.89 & 0.20 & $10^{-7}$ & 95.8 & 96.3 & 96.3 & 96.7 \\
R9 & 1000 & 0.90 & 0.10 & $10^{-4}$ & 96.7 & 96.9 & 96.9 & \textbf{97.3} \\
\bottomrule
\end{tabular}
\vspace{-6mm}
\end{wraptable}

Evaluating  nine configurations in Table~\ref{tab:hps} across all context inputs (36 runs) showed that reservoir capacity is the dominant factor: accuracy increases monotonously with $N$ across all context configurations, while varying other hyperparameters at fixed $N=300$ changes accuracy by at most $0.5$\,pp. 
Context gain diminishes with reservoir capacity: at $N=30$, full context contributes $+4.1$\,pp; 
at $N=1{,}000$, only $+0.60$\,pp. This confirms that contextual features compensate for limited reservoir memory rather than providing independent discriminative signal. 
The best overall result ($97.3\%$ for R9 with \textit{ctx-all}) is reported for completeness.

\subsection{Single-Trial Analysis}

\begin{figure}[t!]
  \centering
  \includegraphics[width=\textwidth]{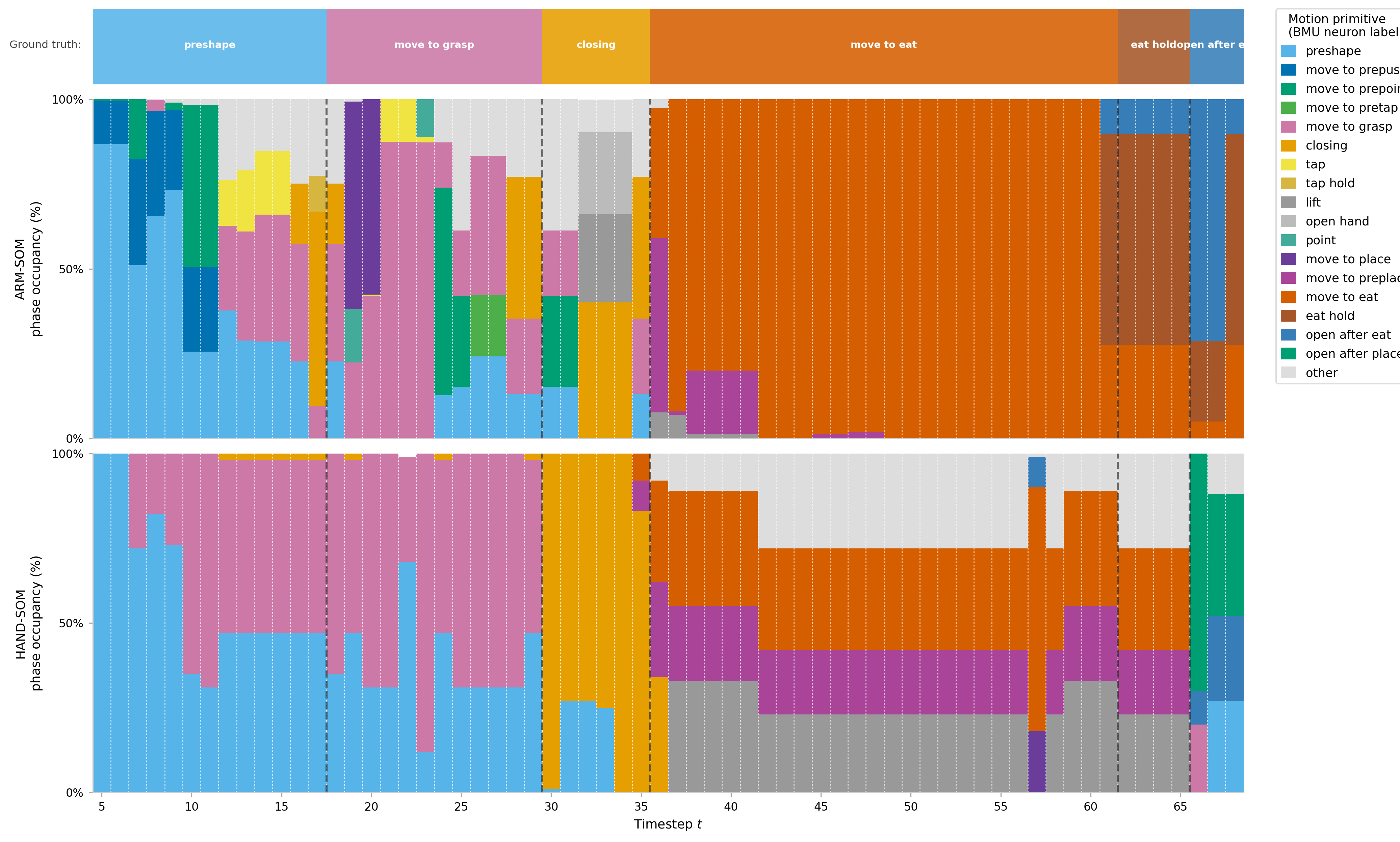}
  \caption{ESN phase recognition timeline for representative \textit{Eat} trial.
    \textit{Top}: ground-truth phase sequence.
    \textit{Middle}: phase distribution of the active A-SOM neuron at each timestep.
    \textit{Bottom}: phase distribution of the active H-SOM neuron.
    The complementary coverage of A-SOM and H-SOM is visible: A-SOM achieves near-pure move-to-eat representation ($t=36$--$61$) while H-SOM provides sharp closing discrimination at $t=30$.
    }
  \label{fig:trial_timeline}
\end{figure}

We traced the activation trajectory of an \textit{Eat} trial and compared per-step predictions of \textit{ctx-no} and \textit{ctx-all} against ground truth (see Figure~\ref{fig:trial_timeline}).
\textit{Ctx-no} has $98.4\%$ accuracy (63/64 steps), with a single error at $t=18$ (the first \textit{move-to-grasp} timestep) because the reservoir had accumulated 13 consecutive preshape steps and required one additional step to transition.
\textit{Ctx-all} achieves $100\%$: the action one-hot (Eat) and the palm-to-object distance at $t=18$ provide a sufficient signal to detect the phase transition one step earlier.

The result also reveals neuron-level ambiguity. A-SOM BMU~$(7,18)$ contains $35\%$ \textit{move-to-grasp}, $23\%$ \textit{preshape}, and $18\%$ \textit{closing} — no dominant phase. 
The ESN resolves this through two complementary mechanisms: the temporal integration of preceding reservoir states, and the simultaneous integration of both A-SOM and H-SOM descriptors. 
At $t=18$, the hand BMU already reflects finger opening -- a signal invisible to  A-SOM alone. Neither SOM provides unambiguous information throughout the trial; together they cover each other's ambiguous regions, and ESN integrates both streams to achieve $98.4\%$.

\section{Conclusion}
\label{sec:discussion}
The results support the hypothesis that self-organised motor primitives provide a useful intermediate representation for MNS-inspired phase recognition. The ctx-no ESN achieved $93.86\%$ phase-recognition accuracy, showing that SOM activation trajectories preserve most phase-discriminative information without world-state context. The small improvement obtained by adding action identity, distance, and contact state suggests that contextual information provides useful refinement, but is not the dominant source of phase-discriminative structure. 
The dual-SOM results confirm that arm and hand kinematics encode complementary aspects of motor behaviour. A-SOM captures the spatial and temporal structure of arm transport, while H-SOM captures hand configuration, supporting separate primitive spaces over a single combined map.

A current limitation is that the model is trained and evaluated on NICO's own actions. Extending it to cross-embodiment observation  will require mirroring mechanisms for handling viewpoint, scale, and embodiment differences.
In our future work, we plan to replace recognition with prediction by training the ESN to anticipate the next SOM activation, enabling primitive prediction and, eventually, closed-loop generation of imagined motor trajectories.

\begin{credits}
 \subsubsection{\ackname}  We thank anonymous reviewers for detailed feedback. This research was supported by project APVV-21-0105.
\end{credits}

\bibliographystyle{splncs04}
\bibliography{references}

\end{document}